\documentclass[11pt]{article}

\usepackage[preprint]{acl}

\usepackage{times}
\usepackage{latexsym}

\usepackage[T1]{fontenc}
\usepackage[utf8]{inputenc}
\usepackage{microtype}
\usepackage{inconsolata}

\usepackage{amsmath}
\usepackage{booktabs}
\usepackage{multirow}
\usepackage{tabularx}
\usepackage{graphicx,enumitem}

\usepackage{comment}

\title{Topic-to-Timestamp Alignment by Constrained Evidence Selection}

\author{
  Zeynep Yılbırt \\
  TU Dresden, Germany \\
  SpeechMind GmbH, Germany \\
  \texttt{zeynep.yilbirt@mailbox.tu-dresden.de}
  \And
  Marina Litvak \\
  Shamoon College of Engineering, Israel \\
  TU Dresden, Germany \\
  \texttt{litvak.marina@gmail.com}
  \AND
  Michael Färber \\
  TU Dresden, Germany \\
  ScaDS.AI Dresden/Leipzig, Germany \\
  \texttt{michael.faerber@tu-dresden.de}
}

\begin{document}
\maketitle

\begin{abstract}
Meeting archives are difficult to search when users remember what was discussed but not when. We study \emph{topic-to-timestamp alignment}: given a natural-language topic and a timestamped meeting transcript, the goal is to return the time at which the topic is discussed. A standard RAG setup can retrieve relevant transcript excerpts, but still asks the language model to generate a timestamp, which can produce unsupported or invalid timecodes. We therefore recast timestamp prediction as constrained temporal candidate selection: the system retrieves timestamped transcript chunks, and the model selects the candidate that best grounds the topic instead of generating a timecode. On 420 topic-timestamp queries from 200 municipal meeting transcripts, this increases Recall@5 from 31.9\% to 50.0\%, reduces MAE from 837.0s to 761.0s with Mistral-7B-Instruct, and increases the number of parseable outputs from 373 to 419 of 420 queries. The results suggest that temporal grounding in long transcripts depends strongly on retrieval quality and output design, not only on the choice of the language model.
\end{abstract}

\section{Introduction}

Long meeting recordings are easy to archive but difficult to search. 
A user may know that a budget issue, project risk, or personnel decision was discussed, but not when it occurred. 
This creates a simple retrieval problem: given a natural-language topic, find the moment in a long transcript where the topic is discussed.
Recent meeting datasets have made long transcripts increasingly accessible for summarization and analysis \citep{hu2023meetingbank}, but temporal localization from open-ended topic queries remains less studied.

Large language models (LLMs) and retrieval-augmented generation (RAG) provide a natural interface for such queries \citep{lewis2020retrieval}. 
Yet temporal grounding exposes a limitation that standard RAG does not remove. 
When asked to return a timestamp, an LLM may produce a plausible time even when the retrieved evidence does not support it. 
We refer to this as \emph{temporal hallucination}: producing a valid-looking timestamp that is not grounded in the transcript, related to broader hallucination problems in neural text generation \citep{ji2023survey}.

We study this problem as \emph{topic-to-timestamp alignment}. 
Given a topic query and a timestamped meeting transcript, the task is to identify the transcript segment that grounds the query and return its corresponding time. 
Unlike standard document retrieval, the output is not only a relevant text span but a temporal location. 
Unlike free-form question answering, a small numerical error can make the answer practically useless: a timestamp that is several minutes off sends the user to the wrong part of the meeting.

We recast timestamp prediction as constrained temporal candidate selection. 
Rather than asking the model to generate a timecode, we first construct a small set of temporally coherent candidate chunks through retrieval. 
The language model then performs only the semantic decision: selecting the candidate that best grounds the topic. 
The final timestamp follows from the selected candidate. 
This separates two operations that standard RAG conflates: finding the relevant discussion and generating a numerical timecode.

We implement this reformulation in a training-free, text-only RAG pipeline. 
The pipeline combines overlapping temporal windows, hybrid sparse-dense retrieval with Reciprocal Rank Fusion \citep{cormack2009rrf}, and constrained Chunk-ID selection. 
For evaluation, we construct a benchmark of municipal meeting transcripts with manually curated topic-timestamp queries.

Our experiments evaluate this reformulation on 420 topic-timestamp queries from 200 municipal meeting transcripts, with additional checks on the full 600-query candidate set. 
The main comparison shows that retrieval and grounding design matter more than model scale alone. 
Replacing Standard RAG with our retrieval-and-selection pipeline raises Recall@5 from 31.9\% to 50.0\%, increases parseable outputs for Mistral-7B-Instruct from 373 to 419 of 420 queries, and reduces MAE for both Mistral-7B-Instruct and GPT-5.4. 
The strongest setting, GPT-5.4 with our pipeline, achieves the best Exact@30s and MAE, while Exact@30s remains substantially below Recall@5. 
This indicates that fine-grained boundary localization remains difficult even when retrieval coverage improves.

Overall, our contributions are:
\begin{itemize}[itemsep=1pt, topsep=2pt, parsep=0pt, partopsep=0pt]
    \item We introduce \emph{topic-to-timestamp alignment} as a temporal grounding task for long meeting transcripts.
    \item We provide a benchmark of municipal meeting transcripts with manually curated topic-timestamp annotations.\footnote{Code \& data at \url{https://github.com/faerber-lab/topic-timestamp-alignment}} 
    \item We propose a training-free RAG pipeline that replaces free-form timestamp generation with constrained Chunk-ID selection.
    \item We show that retrieval and output design improve temporal grounding more reliably than scaling the language model alone.
\end{itemize}

\section{Related Work}

\paragraph{Meeting understanding.}
Meeting understanding has mainly been studied through segmentation, summarization, and query-focused access. 
Topic segmentation identifies coherent topical boundaries in multiparty dialogue \citep{hsueh2006automatic}, while QMSum asks systems to locate relevant spans and summarize them in response to user queries \citep{zhong2021qmsum}. 
MeetingBank provides a large 
benchmark of city council meetings with transcripts, agendas, minutes, and aligned summaries \citep{hu2023meetingbank}. 
These works support structured access to meetings, but they 
target segmentation or summarization rather than returning the temporal location of a user-specified topic.

\paragraph{Temporal localization and retrieval.}
Temporal localization is central to video retrieval, where systems map natural-language queries to moments in untrimmed videos \citep{paul2021text}. 
Recent multimodal systems combine visual, audio, and textual signals for video understanding and retrieval \citep{zhang2023video,rahman2025ai}. 
Our setting is conceptually related because it also grounds natural-language queries in time, but differs in focusing on timestamped transcripts only. 
This makes the task lighter-weight and easier to integrate into existing meeting archives, while removing visual and acoustic cues that multimodal systems can exploit.

\paragraph{Retrieval-augmented and constrained generation.}
Retrieval augmented generation (RAG) incorporates external evidence into generation and is widely used for long-context tasks \citep{lewis2020retrieval,izacard2021leveraging}. 
For long documents, retrieval quality depends on how texts are segmented, indexed, and ranked. Hybrid retrieval with Reciprocal Rank Fusion offers a way to combine sparse and dense rankings without score calibration \citep{cormack2009rrf}. 
However, standard RAG still often relies on free-form generation over retrieved evidence, which is brittle for numerical timestamp prediction. 
Our approach instead treats topic-to-timestamp alignment as constrained evidence selection over retrieved temporal candidates, avoiding free-form timecode generation. 
This connects to evidence-based prediction benchmarks such as ERASER \citep{deyoung2020eraser} and in-context selection, where models choose among structured alternatives provided in the prompt \citep{dong2024survey}.

\section{Method}
\label{sec:method}

We frame topic-to-timestamp alignment as a constrained evidence selection over temporally indexed transcript chunks. 
Given a timestamped transcript $T$ and a natural-language topic query $q$, the goal is to return the timestamp of the transcript segment that best grounds $q$. 
Instead of asking an LLM to generate a timestamp directly, our method first retrieves candidate chunks and then asks the model to select one of their discrete identifiers. 
The selected identifier determines the returned timestamp.

\paragraph{Temporal candidates.}
We segment each transcript into overlapping temporal windows. 
Each window preserves local conversational context and acts as one candidate unit for retrieval and grounding. 
In the main configuration, we use 90-second windows with a 45-second overlap, selected based on the development ablation in Appendix~\ref{app:ablations}. 
Each candidate chunk stores its transcript text, start time, end time, and a synthetic identifier such as \texttt{[C01]}.

\paragraph{Hybrid retrieval.}
For each topic query, we retrieve candidate chunks using both dense and sparse signals. 
Dense retrieval with \texttt{BAAI/bge-large-en-v1.5} captures semantic matches and paraphrased topic descriptions, while BM25 preserves exact lexical evidence such as names, agenda terms, and domain-specific expressions. 
This distinction is important in meeting transcripts, where the same topic may be referred to indirectly, but agenda items and speaker names often remain lexical anchors. 
Because dense and sparse scores are not directly comparable, we fuse ranked lists rather than raw scores. 
We use Reciprocal Rank Fusion (RRF) \citep{cormack2009rrf}, a rank-level method that requires no score calibration:
\begin{equation*}
    \mathrm{RRF}(d) =
    \frac{1}{k + \mathrm{rank}_{\mathrm{dense}}(d)}
    +
    \frac{1}{k + \mathrm{rank}_{\mathrm{sparse}}(d)} ,
\end{equation*}
where $d$ is a candidate chunk and $k$ is set to 60. 
This favors chunks that are highly ranked by either retriever, while giving additional weight to chunks supported by both. 
The top-ranked chunks form the evidence set shown to the LLM.

\paragraph{Constrained Chunk-ID selection.}
The final grounding step turns timestamp prediction into candidate selection. 
Rather than prompting the model to produce a raw timestamp, we present the retrieved candidates with their Chunk IDs and ask the model to output only the ID of the candidate that best supports the topic query. 
The selected ID is then resolved to the candidate's start timestamp. 
This prevents the model from inventing unsupported timecodes and separates semantic grounding from numerical timestamp generation.

This design changes the prediction space of the model. 
The model no longer has to generate an arbitrary numerical value, but chooses among a small set of retrieved temporal candidates. 
As a result, retrieval errors and selection errors become separable: if the correct chunk is not retrieved, the failure is a retrieval failure; if it is retrieved but not selected, the failure is a grounding failure.

\section{Experimental Setup}
\label{sec:setup}

\paragraph{Dataset.}
We evaluate topic-to-timestamp alignment on municipal meeting transcripts. 
Existing meeting benchmarks are adjacent but do not directly evaluate this task: QMSum \citep{zhong2021qmsum} targets query-based meeting summarization, MeetingBank \citep{hu2023meetingbank} focuses on city council meeting summarization and minute alignment, and video moment-localization datasets \citep{paul2021text} are designed for multimodal grounding rather than text-only transcript grounding. 
We therefore construct a task-specific benchmark from 200 public municipal meeting recordings. 
After automatic speech recognition and speaker diarization, we curate 600 topic--timestamp pairs and apply a predefined procedural filter that removes repetitive opening and closing items such as roll call, pledges, invocations, and adjournments. 
This yields a 420-query benchmark used for all main experiments. 
The filtering is applied before model evaluation and depends only on heading categories, not on model predictions. 
Dataset construction details and the full 600-query comparison are provided in Appendix~\ref{app:dataset}.

\paragraph{Systems.} %
We compare our constrained grounding pipeline against a standard RAG baseline that tests whether temporal grounding can be solved by retrieval plus direct timestamp generation alone. 
The baseline uses dense retrieval over 30-second transcript windows and prompts the model to generate a timestamp from the retrieved context. 
Our system uses 90-second overlapping windows, hybrid BM25-dense retrieval with RRF, and constrained Chunk-ID selection. 
Both systems retrieve from the same source transcript and return one timestamp per topic query.

\begin{table*}[!t]
\centering
\small
\begin{tabular}{llcccc}
\toprule
\textbf{Setting} & \textbf{Model} & \textbf{Recall@5} & \textbf{Exact@30s} & \textbf{MAE $\downarrow$} & \textbf{Parsed} \\
\midrule
\multirow{4}{*}{Standard RAG}
& Mistral-7B-Instruct & 31.9\% & 28.8\% & 837.0s & 373/420 \\
& GPT-5.1 & 31.9\% & 28.6\% & 1115.1s & 420/420 \\
& GPT-5.2 & 31.9\% & 28.6\% & 1080.7s & 418/420 \\
& GPT-5.4 & 31.9\% & 29.3\% & 1118.3s & 406/420 \\
\midrule
\multirow{2}{*}{Ours}
& Mistral-7B-Instruct & \textbf{50.0\%} & 30.0\% & 761.0s & 419/420 \\
& GPT-5.4 & \textbf{50.0\%} & \textbf{34.3\%} & \textbf{743.3s} & \textbf{420/420} \\
\bottomrule
\end{tabular}
\caption{
Main results on our 420-query topic--timestamp benchmark. 
\emph{Standard RAG} denotes dense retrieval with direct timestamp generation; 
\emph{Ours} denotes the proposed retrieval-and-selection pipeline. 
Recall@5 is retrieval-level and therefore identical across models under the same setting.
}
\label{tab:main-results}
\end{table*}

\paragraph{Models.}
We evaluate Mistral-7B-Instruct and several GPT models accessed through the OpenAI API. 
Mistral-7B is used for the main ablations because its open weights improve reproducibility and make retrieval and output-constraint effects easier to isolate. 
The GPT models serve as stronger proprietary comparisons, testing whether model scale alone can compensate for weak retrieval or free-form timestamp generation. 
Exact API model identifiers, decoding settings, access dates, and prompt templates are reported in Appendix~\ref{app:model-config}. 

\paragraph{Metrics.}
We report four metrics. 
\textbf{Recall@5} measures whether the gold timestamp falls within one of the top five retrieved chunks, separating retrieval failure from selection failure. 
\textbf{Exact@30s} is the percentage of predictions within $\pm 30$ seconds of the gold timestamp. 
\textbf{MAE} is the mean absolute timestamp error in seconds and penalizes large temporal deviations. 
\textbf{Parsed} is the number of queries for which the system returns a parseable output. 
Invalid timestamps, empty responses, and unparseable Chunk IDs are counted as unparsed.

\section{Results and Analysis}
\label{sec:results}

Table~\ref{tab:main-results} separates model effects from system-design effects. 
The upper block keeps the baseline retrieval and grounding setup fixed and varies only the model. 
Here, GPT models produce more parseable outputs than Mistral-7B-Instruct, but do not improve Exact@30s and yield higher MAE, suggesting that stronger instruction following alone does not compensate for weak temporal retrieval and free-form timestamp generation. 
The lower block reports our proposed pipeline, which replaces short-window dense retrieval with Hybrid RRF over longer temporal windows and replaces free-form timestamp generation with Chunk-ID selection. 
This raises Recall@5 from 31.9\% to 50.0\% and reduces temporal error for both Mistral-7B-Instruct and GPT-5.4.

Table~\ref{tab:full-vs-filtered} in the Appendix compares the filtered benchmark with the full 600-query candidate set. 
Using the same Mistral-7B-Instruct pipeline, the full set reaches higher scores: 56.3\% Recall@5, 37.3\% Exact@30s, and 632.0s MAE. 
This suggests that filtering removes easier, repetitive cases rather than making the benchmark easier.

The following analyses clarify where the gains come from and where the task remains difficult.

\paragraph{Retrieval is the bottleneck.}
The largest improvement is candidate coverage. 
If the relevant segment is not retrieved, no downstream model can select the correct timestamp. 
The retrieval ablation in Table~\ref{tab:retrieval-ablation} shows the same pattern: moving from dense retrieval with 30-second windows and 15-second overlap to Hybrid RRF with 90-second windows and 45-second overlap raises Recall@5 from 31.9\% to 50.0\%. 
Topic-to-timestamp alignment should therefore first be treated as evidence retrieval and only then as generation.

\paragraph{Constrained selection improves robustness.}
Using the same Hybrid RRF retrieval setting, free-form timestamp generation with Mistral-7B-Instruct obtains 27.6\% Exact@30s, 1048.7s MAE, and parseable outputs for 387 of 420 queries. 
Replacing timestamp generation with Chunk-ID selection improves these values to 30.0\%, 761.0s, and 419 of 420 queries, as shown in Table~\ref{tab:freeform-vs-chunkid}. 
The same pattern appears in tail errors: the 95th percentile of absolute error decreases from 3962.2s to 3313.9s for Mistral-7B-Instruct and from 4966.0s to 3173.4s for GPT-5.4 (Table~\ref{tab:tail-errors}). 
The main effect is therefore improved robustness: fewer invalid outputs and fewer large temporal errors.


\paragraph{Fine-grained boundaries remain difficult.}
The gains are larger for retrieval coverage and temporal-error reduction than for exact boundary accuracy. 
In the strongest setting, Recall@5 reaches 50.0\%, while Exact@30s remains 34.3\%. 
This gap suggests that the system often retrieves a relevant temporal region without selecting the exact annotated start time. 
A retrieved chunk may contain the correct discussion but start before the annotated topic boundary, and many topics emerge gradually across several turns. 
The method therefore addresses temporal hallucination and large-error reduction, but not fine-grained boundary detection.

\section{Conclusion}
\label{sec:conclusion}

We introduced topic-to-timestamp alignment for long meeting transcripts and showed that temporal grounding should not be treated as free-form timestamp generation. 
By retrieving temporal candidates and constraining the model to select evidence, our pipeline makes retrieval coverage explicit and reduces invalid or unsupported timecodes. 
Overall, reliable grounding in timestamped transcripts depends less on scaling the language model alone than on the interface between retrieval, evidence selection, and temporal metadata.



\section*{Limitations}

The benchmark is based on municipal meeting transcripts and currently relies on single-annotator curation. 
The results therefore support claims about text-only temporal grounding in this domain, but should be validated on other meeting types, languages, and with multiple annotators.

The pipeline also depends on transcript quality. 
ASR or diarization errors can shift, remove, or obscure the evidence before retrieval, and our evaluation does not separately isolate these upstream errors from retrieval and selection failures.

Finally, Chunk-ID selection reduces invalid and unsupported timecodes, but it does not solve boundary ambiguity. 
Many topics emerge gradually across several turns, so a single annotated start time can be sharper than the underlying discussion. 
The GPT results are also tied to the accessed API snapshots. We therefore use Mistral-7B-Instruct for the main ablations and report exact model identifiers in Appendix~\ref{app:model-config}.



\bibliography{custom}

\clearpage
\appendix

\section{Dataset Construction}
\label{app:dataset}

\paragraph{Transcript construction.}
We collected 200 public municipal meeting recordings. 
Audio was converted to 16 kHz mono using FFmpeg. 
We used Whisper-medium for ASR and \texttt{pyannote/speaker-diarization} for speaker segmentation. 
Word-level ASR outputs were assigned to speaker segments by temporal overlap. 
The resulting transcripts contain timestamped speaker turns, e.g.:
\begin{quote}
\small
\verb|[00:15:32] SPEAKER_01: We now discuss|\\
\verb|the budget amendment...|
\end{quote}

\paragraph{Annotation.}
The candidate set contains 600 topic--timestamp pairs. 
Each instance consists of a source transcript, a topic heading, and a gold timestamp marking the start of the corresponding discussion. 
When public video descriptions provided timestamps, these were used as initial boundaries and checked against the transcript. 
Missing boundaries were identified manually from the video and transcript. 
The dataset was curated by a single annotator; we therefore do not report inter-annotator agreement.

\paragraph{Procedural filtering.}
The full candidate set contains repetitive procedural headings, including call-to-order, roll call, pledge, invocation, opening remarks, and adjournment. 
We remove these categories before evaluation, yielding the 420-query benchmark used in the main paper. 
The filtering is based only on heading categories. 
Table~\ref{tab:full-vs-filtered} reports the full candidate set and the filtered benchmark under the same pipeline.

\begin{table*}[t]
\centering
\small
\begin{tabular}{lccccc}
\toprule
\textbf{Set} & \textbf{Queries} & \textbf{Recall@5} & \textbf{Exact@30s} & \textbf{MAE $\downarrow$} & \textbf{Parsed} \\
\midrule
Full candidate set & 600 & \textbf{56.3\%} & \textbf{37.3\%} & \textbf{632.0s} & 600/600 \\
Filtered benchmark & 420 & 50.0\% & 30.0\% & 761.0s & 419/420 \\
\bottomrule
\end{tabular}
\caption{Full candidate set and filtered benchmark using the same Mistral-7B-Instruct pipeline.}
\label{tab:full-vs-filtered}
\end{table*}

\section{Model and Prompt Configuration}
\label{app:model-config}

\paragraph{Models.}
The open-source model is \texttt{mistral:7b-instruct}, served through Ollama. 
The GPT model identifiers are:

\begin{itemize}
    \item GPT-5.1: \texttt{gpt-5.1-2025-11-13}
    \item GPT-5.2: \texttt{gpt-5.2-2025-12-11}
    \item GPT-5.4: \texttt{gpt-5.4-2026-03-05}
\end{itemize}

All GPT models were accessed in May 2026. 
Decoding settings and parsing rules are included in the evaluation configuration.

\paragraph{Timestamp-generation prompt.}
\begin{quote}
\small
\textbf{Task:} Given the topic and transcript excerpts, identify when the topic is discussed.

\textbf{Topic:} \{topic\}

\textbf{Transcript excerpts:} \{retrieved\_context\}

Return only one timestamp in \texttt{HH:MM:SS} format.
\end{quote}

\paragraph{Chunk-ID prompt.}
\begin{quote}
\small
\textbf{Task:} Select the candidate chunk that best supports the topic.

\textbf{Topic:} \{topic\}

\textbf{Candidates:}

\texttt{[C01]} \{chunk\_text\}

\texttt{[C02]} \{chunk\_text\}

\ldots

Return only the Chunk ID, e.g., \texttt{[C03]}.
\end{quote}

\paragraph{Parsing.}
For timestamp generation, we parse \texttt{HH:MM:SS}, \texttt{MM:SS}, and unambiguous natural-language times. 
For Chunk-ID selection, we parse the first valid candidate ID. 
Empty or unparseable outputs count as unparsed.

\section{Additional Ablations}
\label{app:ablations}

Table~\ref{tab:retrieval-ablation} reports the retrieval and window-size ablation. 
Table~\ref{tab:freeform-vs-chunkid} isolates the effect of the final grounding format under the same retrieval setup.

\begin{table*}[t]
\centering
\small
\begin{tabular}{lccc}
\toprule
\textbf{Retrieval setting} & \textbf{Recall@5} & \textbf{Exact@30s} & \textbf{MAE $\downarrow$} \\
\midrule
Dense + 30s/15s windows & 31.9\% & 28.8\% & 837.0s \\
Dense + 90s/45s windows & 41.9\% & 28.8\% & 818.0s \\
Hybrid RRF + 30s/15s windows & 39.0\% & \textbf{30.5\%} & 834.4s \\
Hybrid RRF + 90s/45s windows & \textbf{50.0\%} & 30.0\% & \textbf{761.0s} \\
\bottomrule
\end{tabular}
\caption{Retrieval and window-size ablation with Mistral-7B-Instruct.}
\label{tab:retrieval-ablation}
\end{table*}

\begin{table*}[t]
\centering
\small
\begin{tabular}{lcccc}
\toprule
\textbf{Grounding strategy} & \textbf{Recall@5} & \textbf{Exact@30s} & \textbf{MAE $\downarrow$} & \textbf{Parsed} \\
\midrule
Free-form timestamp generation & 50.0\% & 27.6\% & 1048.7s & 387/420 \\
Chunk-ID selection & 50.0\% & \textbf{30.0\%} & \textbf{761.0s} & \textbf{419/420} \\
\bottomrule
\end{tabular}
\caption{Free-form timestamp generation and constrained Chunk-ID selection under the same 90s/45s Hybrid RRF retrieval setup.}
\label{tab:freeform-vs-chunkid}
\end{table*}

\section{Error Analysis}
\label{app:error-analysis}

Table~\ref{tab:tail-errors} reports absolute-error statistics for the baseline and proposed systems.

\begin{table}[t]
\centering
\small
\begin{tabular}{lcc}
\toprule
\textbf{System} & \textbf{Median AE} & \textbf{P95 AE} \\
\midrule
Mistral baseline & 150.0s & 3962.2s \\
Mistral proposed & 204.0s & 3313.9s \\
GPT-5.4 baseline & -- & 4966.0s \\
GPT-5.4 proposed & -- & 3173.4s \\
\bottomrule
\end{tabular}
\caption{Absolute-error statistics on the 420-query benchmark.}
\label{tab:tail-errors}
\end{table}

\paragraph{Failure categories.}
We group errors into three categories: (i) the relevant segment is not retrieved, (ii) the selected chunk is topically related but temporally offset, and (iii) the output is invalid or unparseable. 
The proposed system reduces invalid outputs and large outlier errors, but does not remove boundary-offset errors.

\section{Additional Implementation Details}
In the reported hybrid setting, candidates are retrieved with dense retrieval and BM25 using multiple query variants, fused with Reciprocal Rank Fusion, and lightly re-ranked using heading-term overlap and an early-section temporal prior when applicable. These steps only use the query text and the information retrieved from the candidates, not the gold topic timestamps or the model outputs. For the reported Chunk-ID experiments, deterministic fallback extraction is disabled; the model selects a candidate ID, which is then mapped to the selected chunk start time.

\end{document}